\documentclass[lettersize,conference]{IEEEtran}
\usepackage{amsmath,amsfonts}
\usepackage{algorithmic}
\usepackage{algorithm}
\usepackage{array}
\usepackage[caption=false,font=normalsize,labelfont=sf,textfont=sf]{subfig}
\usepackage{textcomp}
\usepackage{stfloats}
\usepackage{url}
\usepackage{verbatim}
\usepackage{graphicx}
\usepackage{tikz}
\graphicspath{{im/}}
\usepackage{cite}
\usepackage{xcolor}

\begin{document}

\title{Find the Differences: Differential Morphing Attack Detection vs Face Recognition}

\author{
	\IEEEauthorblockN{Una M. Kelly\IEEEauthorrefmark{1}\IEEEauthorrefmark{2},
		Luuk J. Spreeuwers\IEEEauthorrefmark{1}, and
		Raymond N.J. Veldhuis\IEEEauthorrefmark{1}\IEEEauthorrefmark{3}}
	
	\IEEEauthorblockA{\IEEEauthorrefmark{1}Data Management and Biometrics, University of Twente, Enschede, the Netherlands\\}
	\IEEEauthorblockA{\IEEEauthorrefmark{2}Machine Learning and Data Engineering, University of Münster, Münster, Germany\\}
	\IEEEauthorblockA{\IEEEauthorrefmark{3}Department of Information Security and Communication Technology,\\
		Norwegian University of Science and Technology, Gj{\o}vik, Norway\\
		Email: ukelly@uni-muenster.de, l.j.spreeuwers@utwente.nl, r.n.j.veldhuis@utwente.nl}
}

\IEEEpubid{0000--0000/00\$00.00~\copyright~2021 IEEE}

\maketitle

\begin{abstract}
	Morphing is a challenge to face recognition (FR) for which several morphing attack detection solutions have been proposed.  We argue that face recognition and differential morphing attack detection (D-MAD) in principle perform very similar tasks, which we support by comparing an FR system with two existing D-MAD approaches. We also show that currently used decision thresholds inherently lead to FR systems being vulnerable to morphing attacks and that this explains the trade-off between performance on normal images and vulnerability to morphing attacks. We propose using FR systems that are already in place for morphing detection and introduce a new evaluation threshold that guarantees an upper limit to the vulnerability to morphing attacks - even of unknown types. 
\end{abstract}

\begin{IEEEkeywords}
Face recognition, morphing attack detection.
\end{IEEEkeywords}

\section{Introduction}

Two images of different persons can be mixed to create a \textit{morph}. In many cases when a Face Recognition (FR) system compares the morph with an image of the first person it is accepted as a match, but also when it compares the morph with an image of the second person \cite{FFM14,MorphingSurvey21}. If either person were to apply for an ID document using such a morph, then the other person could travel under their name, possibly avoiding travel restrictions. The first method proposed to create morphs was based on landmarks \cite{FFM14}. Since then, other methods have been proposed, such as GAN \cite{MIPGAN,kelly2023worst} or Diffusion models \cite{MorDiff}.

\begin{figure}[h]
		\centering
		\subfloat[\label{1a}]{
		\begin{tikzpicture}
		\node[inner sep=0pt] (whitehead) at (0,0)
		{\includegraphics[width=0.35\paperwidth]{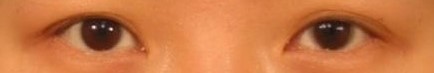}};
		\draw[-,red] (1.5,0.4) to (2.5,0.4) to (2.5,-0.4) to (1.5,-0.4) to (1.5,0.4);
		\end{tikzpicture}}
	
		\subfloat[\label{1a}]{
		\includegraphics[width=0.1\paperwidth]{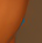}}
		\hspace{0.5cm}
		\subfloat[\label{1a}]{
		\includegraphics[width=0.1\paperwidth]{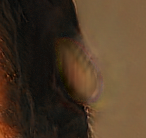}}

		\caption{\label{fig:artefacts} Examples of artefacts in different morphs. a) Ghosting artefact in a morph due to slightly inaccurately selected landmarks, causing a slight shadow around the right pupil. b) Artefact in a Diffusion morph. c) Artefact in a GAN-based morph.}
	\end{figure}

Combining two images to make a morph can lead to visible traces or artefact, for example, when landmarks were not selected correctly. Landmark morphs also often have a smoother texture than normal images, due to the warping and averaging during morphing. 
Deep-learning-based morphs can also contain artefacts, see Fig. \ref{fig:artefacts}. These traces can be used to develop morphing attack detection (MAD) tools.

However, in countries that allow someone applying for an identity document to provide their own printed passport photo, the images are printed and subsequently scanned before being stored in an electronic Machine Readable Travel Document (eMRTD). During this process morphing traces may be masked. Morphing traces and artefacts can also be masked by applying post-processing techniques, leading to low cross-dataset performance \cite{spreeuwers2018towards}. 

\IEEEpubidadjcol
We think that there is a distinct possibility that carefully made morphs do not contain	enough artefacts - such as blurriness, ghosting or other artefacts, which can all be masked by post-processing and/or printing\&scanning - to allow morphing detection to distinguish them from real photos. That means we cannot rely on single-image morphing attack detection (\mbox{S-MAD}) to detect high quality morphs. This is supported by benchmarks such as \cite{NIST_frvt} that show that S-MAD methods have a harder time distinguishing between genuine and morphed images than differential morphing attack detection (D-MAD) methods, where a trusted probe image is available to compare with the suspected morph. S-MAD methods may play a useful role in detecting some types of morph, such as ones that contain some morphing artefacts and have not been post-processed, but in other cases it is likely that D-MAD methods are more appropriate. For this reason we limit our scope to D-MAD, using two well-known approaches: morphing detection based on deep face representations \cite{9093905} and demorphing \cite{ferrara2017face}. We evaluate how well an FR system can perform morphing detection, and also evaluate these two D-MAD methods as if they were FR systems. To our knowledge we are the first to consider the performance of D-MAD systems on non-mated pairs.

Deep learning-based FR systems take images and project them onto \emph{embeddings} in a \emph{latent space}. These embeddings are abstract representations of the identity features in a face. Pairs of faces can be compared by calculating (dis)similarity scores using the corresponding embeddings. For existing FR systems and several different morphing approaches, it has been shown that there is a trade-off between the performance on normal data and vulnerability to morphing attacks, i.e. the better an FR system is at distinguishing between mated and non-mated scores, the more vulnerable it is to morphing attacks \cite{GANmorphdetection, kelly2023worst, NIST_frvt}. We introduce a model based on von Mises-Fisher (vMF) distributions for sampling latent embeddings of multiple identities. Simulated embeddings can be used to compute score distributions that help us explain why the performance-vulnerability trade-off exists and illustrate that if FR systems are further improved in the future, the vulnerability to morphing attacks will be further increased, providing a warning against continuing to set evaluation thresholds according to current convention. Instead, we propose an alternative way to select decision thresholds for FR systems that guarantees an upper limit to the vulnerability to \emph{any} type of morphing attack.

In summary, our contributions are as follows:
\begin{itemize}
	\item comparing the morphing detection capabilities of two differential morphing attack detection (D-MAD) schemes with an FR system, which we show can also be used for morphing detection,
	\item introducing a model for sampling latent embeddings of multiple identities and using these to simulate score distributions in the context of morphing attacks,
	\item showing that the trade-off between performance on normal images and vulnerability to morphing attacks is largely due to the way decision thresholds are chosen,
	\item proposing a new decision threshold for FR systems that guarantees an upper bound for the vulnerability to morphing attacks, even for unknown morphing tools.
\end{itemize}

\section{Background}
Several Morphing Attack Detection (MAD) methods have been developed \cite{raja2020morphing}. MAD methods are trained and evaluated with image datasets that contain both genuine and morphed images. Any differences between morphed and genuine images are dependent on the type of images present in the dataset. The performance of MAD methods is influenced by whether or not training and testing sets include (morphed) images from different sources, whether images were post-processed, whether they were printed and subsequently scanned - mimicking the process that printed passport photos undergo before they are stored in an electronic Machine Readable Travel Document~(eMRTD), etc. Finally, how well a morphing detection approach works depends on the algorithms or approaches used to create morphed images, which can concern
\begin{itemize}
	\item the morphing tool used, which can be for example landmark-, GAN- or diffusion-based,
	\item the landmarks used, whether they were hand-selected or automatically detected using a tool like dlib \cite{dlib09},
	\item selection of image pairs for morphing.
\end{itemize} 
Depending on the data an FR system or morphing detection algorithm was trained and tested with, its ability to classify morphs may be overestimated. For example, in \cite{GANmorphdetection} it was shown that detection methods do not perform as well in a cross-dataset setting, i.e. when training with one type of morph and testing with another. For this reason, evaluating with worst-case morphs \cite{wcMorphing,kelly2023worst} is useful, since they provide an upper bound on how similar a morph can be to two other given images, even for currently unknown morphing algorithms.


Evaluating with worst-case morphs is possible for D-MAD methods that rely on FR embeddings. Our experiments in Section \ref{exp_results} show that in practice worst-case morphs are also more challenging than any known type of morph. However, they do not provide an upper bound guarantee like they do for FR systems. For other D-MAD methods such as demorphing we cannot calculate a worst-case distribution.

D-MAD techniques include methods that:
\begin{enumerate}
	\item Classify images by extracting embeddings from images and comparing them \cite{9093905, 9067912, 9011378}.
	\item Compare corresponding landmarks in pairs of images to detect discrepancies caused by changes in the geometry of the face \cite{DBV19,SBG18}. While such methods should not rely on morphing traces, they struggle to differentiate between landmark shifts due to slight pose variation and shifts due to morphing.
	\item Reverse the morphing process by demorphing a suspected morph using an available live probe image \cite{ferrara2017face,ortega2020border, peng2019fd}. Identity disentanglement also tries to recover information about the second identity when given a trusted reference image of the first identity \cite{IdDisent}.
	\item Use combinations of the above, e.g. \cite{9897023, CombIDfeatures}.
\end{enumerate} 

Promising results have been shown by the MAD method introduced in \cite{9093905, NIST_frvt}. This approach directly relies on the face representations generated by deep learning-based FR systems. While FR systems use such representations - also called features, embedding vectors or latent embeddings - to verify whether the identity of two facial images match, in \cite{9093905} they are used to train a support vector machine (SVM) that separates bona fide images from morphing attacks. 

For our demorphing experiments, we use the method based on demorphing in image space. 

\section{Experiments and Results}\label{exp_results}
In this section, we first compare the scores that two D-MAD methods assign to mated, morph and non-mated image pairs and compare the results with FR scores. Second, we visualise embeddings in the latent space. Next, we simulate embeddings and use them to determine mated, non-mated and worst-case morphing distributions. Using these simulated distributions, we explain the vulnerability of FR systems. Fourth, we propose a new metric to evaluate an FR system, which prevents inherent vulnerability to morphing attacks. Finally, we give an overview of all data used in our experiments.

\begin{figure*}[hbt!]
	\begin{center}
		\subfloat[]{
			\includegraphics[height=0.26\textheight]{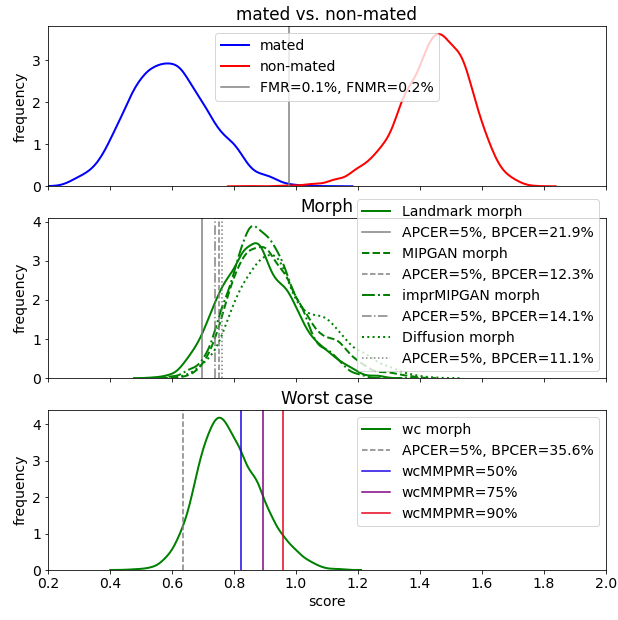}
			\includegraphics[height=0.26\textheight]{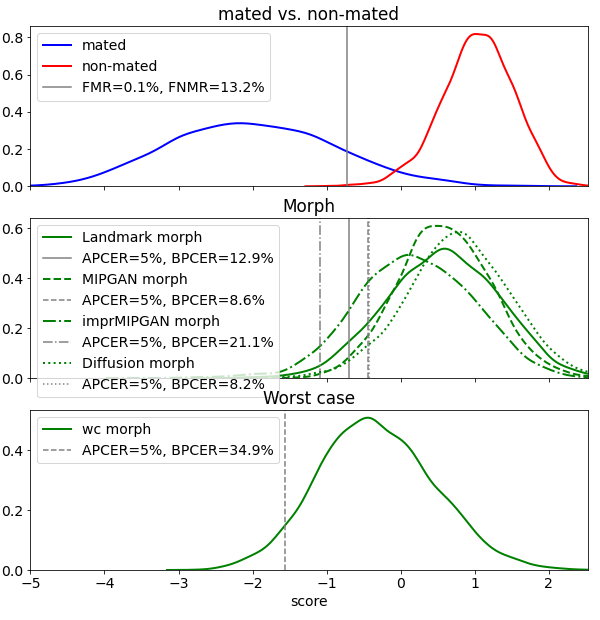}
			\includegraphics[height=0.26\textheight]{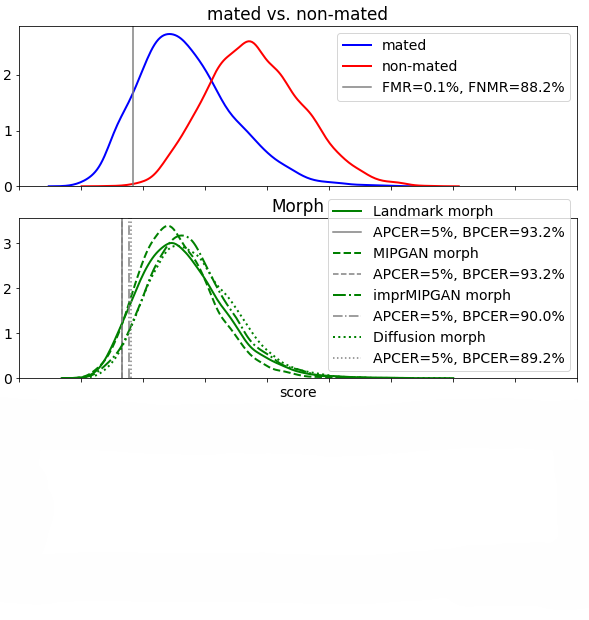}
		}
		\vspace{-0.1cm}
		\subfloat[]{
			\centering
			\includegraphics[height=0.26\textheight]{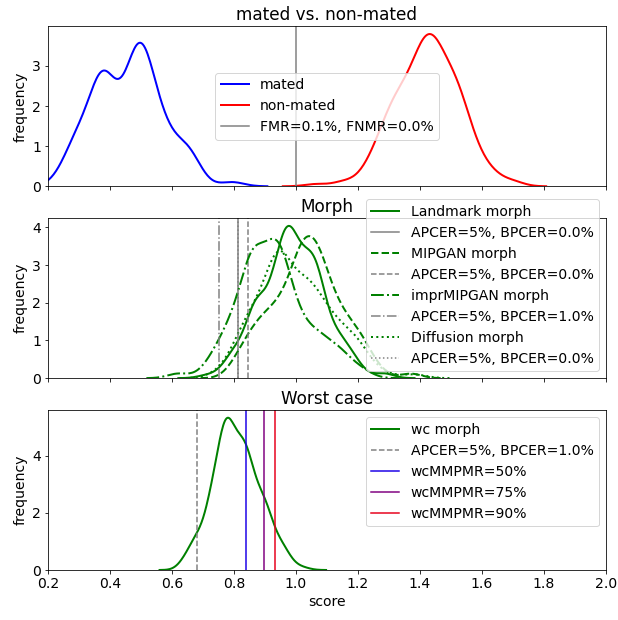}
			\includegraphics[height=0.26\textheight]{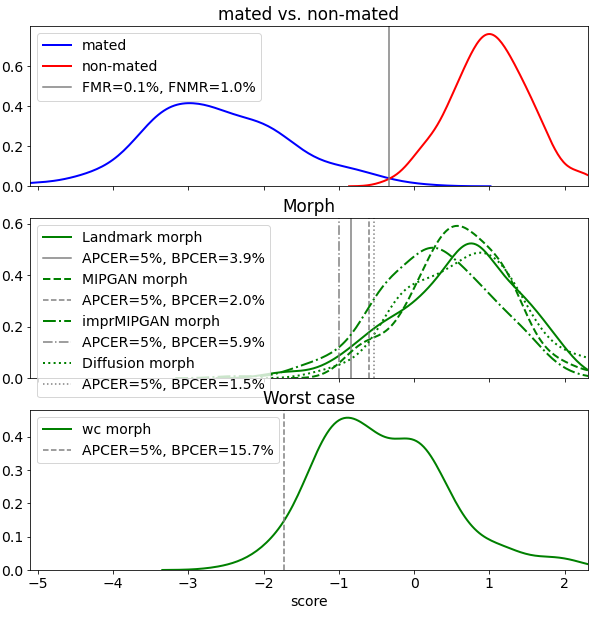}
			\includegraphics[height=0.26\textheight]{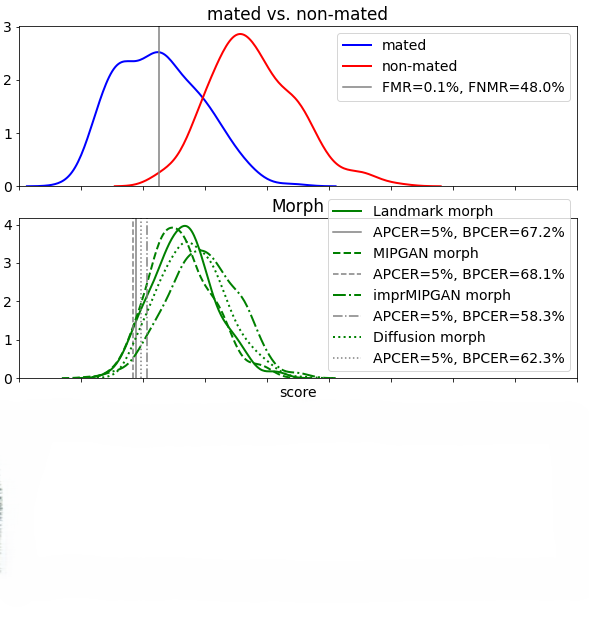}
		}
		\caption{Comparing face recognition (FR) to differential morphing attack detection (D-MAD) using (a) FRGC and (b) FRLL images and morphs. Left: ArcFace scores, middle: D-MAD (based on deep face representations) scores, right: ArcFace scores after demorphing. \label{fig:fr_dmad}}
	\end{center}
\end{figure*}

\subsection{Evaluating morphing attack detection vs. face recognition}\label{sec:MADvsFR}
In this section we evaluate morphing detection under the same conditions as face recognition. We emphasise that the point is not necessarily to evaluate whether FR systems outperform D-MAD or vice-versa, but to illustrate that the two perform very similar tasks. So far, to the best of our knowledge, no-one has examined the behaviour of \mbox{D-MAD} on non-mated image pairs. We show that both \mbox{D-MAD} approaches we examined label practically every single non-mated sample as a morph. This raises the question whether \mbox{D-MAD} performs a significantly different task from FR, since just like FR, D-MAD tries to determine whether two images belong to the same person or not. 

We use the following metrics to evaluate FR and D-MAD:
\begin{itemize}
	\item False Match Rate (FMR): the proportion of non-mated pairs wrongly accepted as a match,
	\item False Non-Match Rate (FNMR): the proportion of mated pairs wrongly rejected,
	\item Attack Presentation Classification Error Rate (APCER): the proportion of attacks that are considered a match,
	\item Bona fide Presentation Classification Error Rate (BPCER): the proportion of genuine pairs that are not accepted,
	\item Mated Morph Presentation Match Rate (MMPMR): the proportion of (morphing) attacks for which probe images of \emph{both} contributing identities are a match.
\end{itemize}
The Morphing Attack Potential (MAP($r,c$)) is the proportion of attacks for which at least $r$ probe images of \emph{both} contributing identities are considered a match for at least $c$ FR systems. The MMPMR is equal to the MAP for $r=1, c=1$. We report only the MMPMR since our experiments with $r>1$ did not reveal new patterns when comparing landmark, (improved) MIPGAN, Diffusion, and worst case morphs. Morphs that are close to worst-case for one FR system are not necessarily worst-case for another, but we can determine worst-case embeddings for each separately to define a wcMAP. 

In Fig. \ref{fig:fr_dmad}, we compare a face recognition system with \mbox{D-MAD}. In both cases, the underlying model is given an image pair as input and outputs a decision score. We can evaluate any model as an FR system by setting the decision threshold at FMR 0.1\%, or as D-MAD by setting the threshold at APCER 5\%. In both settings, D-MAD based on demorphing is outperformed by the FR system, both on FRGC and on FRLL data. Comparisons with the D-MAD based on deep FR embeddings present more mixed results. While it is not at all able to distinguish between morph scores and non-mated scores, on FRGC data it is better at separating mated scores from morph scores, especially those based on landmark morphs, likely because the D-MAD was trained with landmark morphs. It also shows an improved performance on MIPGAN and Diffusion morphs compared to the FR system. However, the worst-case performance remains about the same, and it has actually become more vulnerable to improved MIPGAN morphs. On FRLL data, the D-MAD system is worse at separating any type of morph from mated scores. This could be due to the fact that the FRGC image pairs selected for morphing were already more similar, leading to lower dissimilarity scores for all types of morphs compared to FRLL. This might indicate that this D-MAD approach is most suitable for detecting very challenging morphs. In all cases we see that selecting a stricter decision threshold for the FR system can lead to hugely reduced vulnerability to morphing attacks.

By employing two decision thresholds, an FR system could also be turned into a three-way classifier that distinguishes between mated, morph and non-mated cases. This may have advantages in practice, since it might be useful to be able to distinguish between morphing attacks and impostor attacks.

\subsection{Examining Morphs in Latent Space}\label{sec:latent}
To better explain the concepts introduced in the next two sections, it is helpful to visualise what morphing attacks look like in the latent space of FR systems. For this reason we use tSNE \cite{tSNE} to visualise the 512-dimensional latent space of a pretrained ArcFace model \cite{ArcFace}.

\begin{figure}[h]
	\centering
	\includegraphics[width=0.45\textwidth]{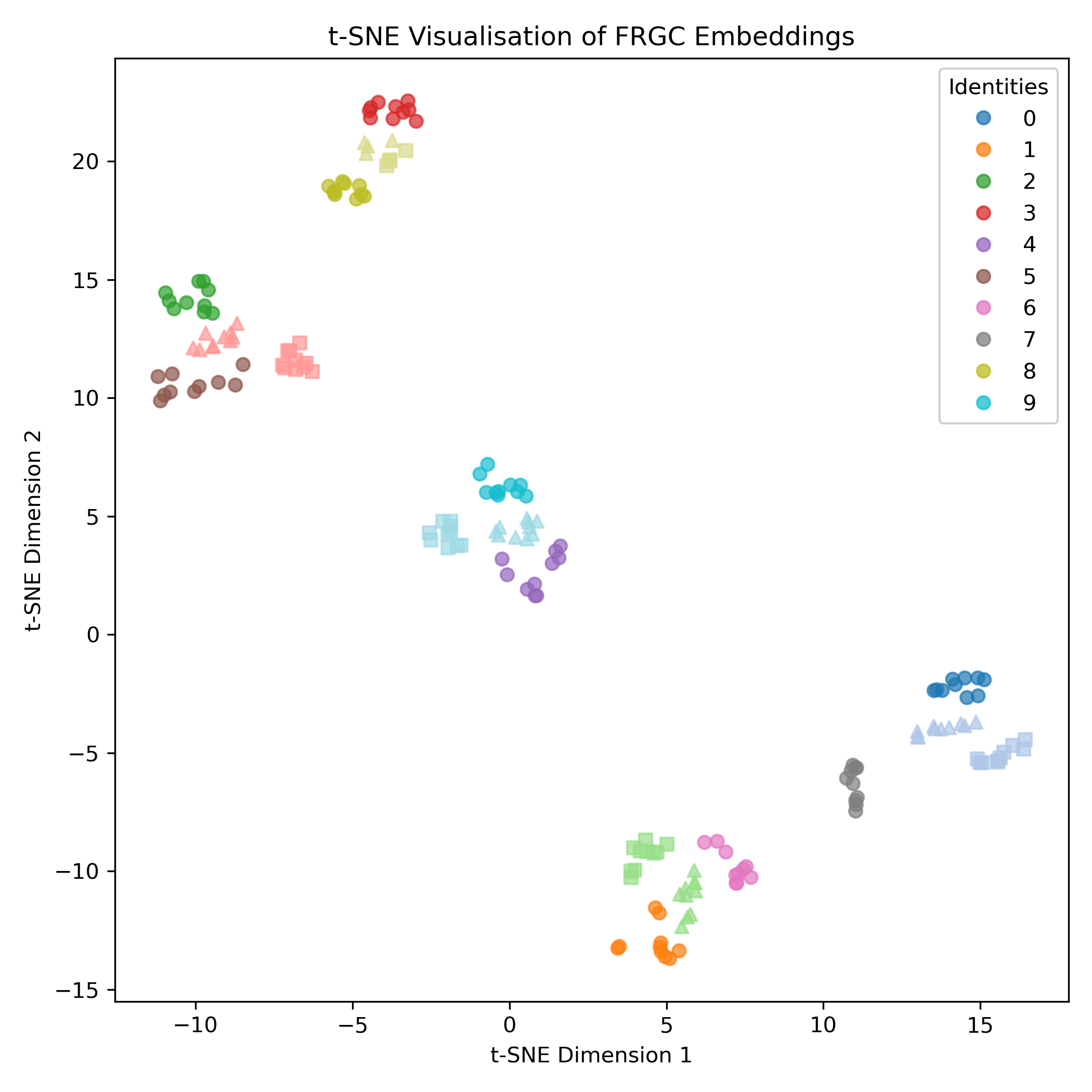}
	\caption{Visualisation with tSNE of the embeddings corresponding to five pairs of normal identities from FRGC \cite{FRGC}. Each pair consists of two identities considered ``similar" by the FR system, see Fig. \ref{fig:SynFace_FRGC_pairs}. Here, we directly applied a pretrained Arcface model \cite{ArcFace} to map images onto embeddings in a 512-dimensional latent space. The closer the embeddings corresponding to morphs (squares) lie to the worst-case embeddings (triangles), the more challenging those morphs are for the FR system. Best seen in colour.\label{fig:visualisation_tSNE}}
	\begin{tikzpicture}[overlay,remember picture]
		\node[inner sep=0pt] (whitehead) at (0.6,8.75)	{\small \text{Identity A}};
		\node[inner sep=0pt] (whitehead) at (2.0,6.5)	{\small \text{Identity B}};
		\node[inner sep=0pt] (whitehead) at (-1.45,6.0)	{\small \text{Landmark morphs}};
		\node[inner sep=0pt] (whitehead) at (-1.45,5.75)	{\small \text{of A and B}};
		\node[inner sep=0pt] (whitehead) at (1.6,7.75)	{\small \text{Worst-case}};
		\node[inner sep=0pt] (whitehead) at (1.6,7.45)	{\small \text{embeddings}};
		
		\draw[black, -] (0.6,8.5) -- (-0.3,7.3);	
		\draw[black, -] (1.3,6.5) -- (0.25,6.4);		
		\draw[black, -] (-1.5,6.2) -- (-0.9,6.6);		
		\draw[black, -] (0.75,7.65) -- (0.0,6.85);	
		
	\end{tikzpicture}
\end{figure}
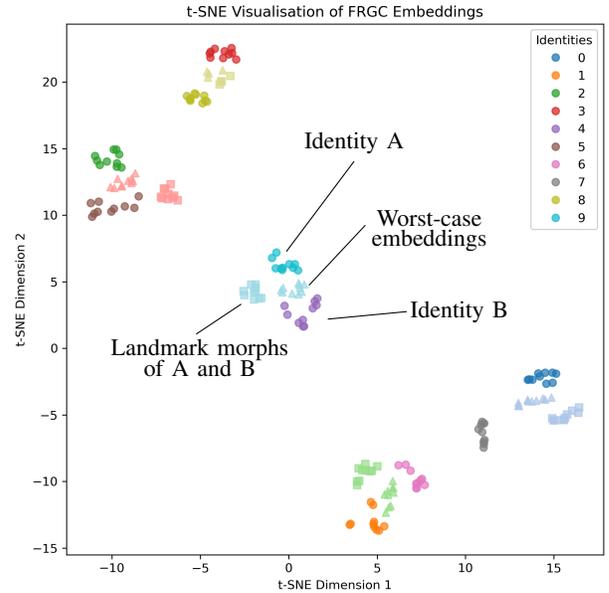

Fig. \ref{fig:visualisation_tSNE} visualises the embeddings corresponding to five pairs of identities from FRGC \cite{FRGC} and the landmark morphs created using images of these identities. For each pair of images used for morphing, we can also compute the worst-case embedding, which represents the most challenging morph based on this pair \cite{wcMorphing}. Furthermore, the most ``vulnerable'' areas in the FR system's latent space are those that lie exactly in between two bona fide class centres. The better a morphing tool is able to generate morphs whose embeddings lie in those vulnerable areas, the more likely it is that the resulting morphing attacks will be successful.

A D-MAD approach that relies on FR embeddings has to learn to separate two possible pairs of samples: either both samples are from the same bona fide cluster or one sample is from a morph cluster and the other from a bona fide cluster. We can see that the FR system itself is able to separate bona fide images and morphs into distinct clusters, as long as a sufficiently strict decision threshold is chosen.

\subsection{Simulating mated and non-mated distributions}\label{sec:sim}
In order to illustrate why improving the performance of FR systems inherently leads to higher vulnerability to morphing attacks and to warn that if FR systems are further improved in the future they will become even more vulnerable to morphing attacks, we simulate embeddings in the FR latent space. To do so, we randomly select $n$ class centres on a hypersphere and use von Mises-Fisher (vMF) distributions to generate samples for each class. We use vMF, since it has proven useful in the context of face recognition \cite{vMF_face, Li2021, Oinar23}. Developing an FR system that significantly improves on the current state of the art would be out of scope for this experiment, but analysing score distributions based on simulated embeddings allows us to examine such a hypothetical FR system.

Given $n$ identities, a good FR system evenly distributes the class centres over the hypersphere in the latent space. Since pairs of high-dimensional vectors will tend to be mutually orthogonal, non-mated scores are centred around $\frac{\pi}{2}\approx 1.57$. That means that the separation of mated and non-mated scores can only be improved by reducing the within-class variation, which is the focus of research such as \cite{CosFace, SphereFace, ArcFace}. The parameter $\kappa$ in vMF can be used to model this variation, where larger $\kappa$ means less within-class variation, i.e. lower mated scores. Since the within-class variation is not the same for each identity, we select $\kappa$ from a random normal distribution. In our simulations, we set $n=250$ and draw 25 samples per identity. The score distributions would not change significantly for different values of $n$, due to the high dimensionality of the latent space, as is shown in e.g. \cite{ArcFace}.

\begin{figure}[h]
	\centering
	\includegraphics[width=0.45\textwidth]{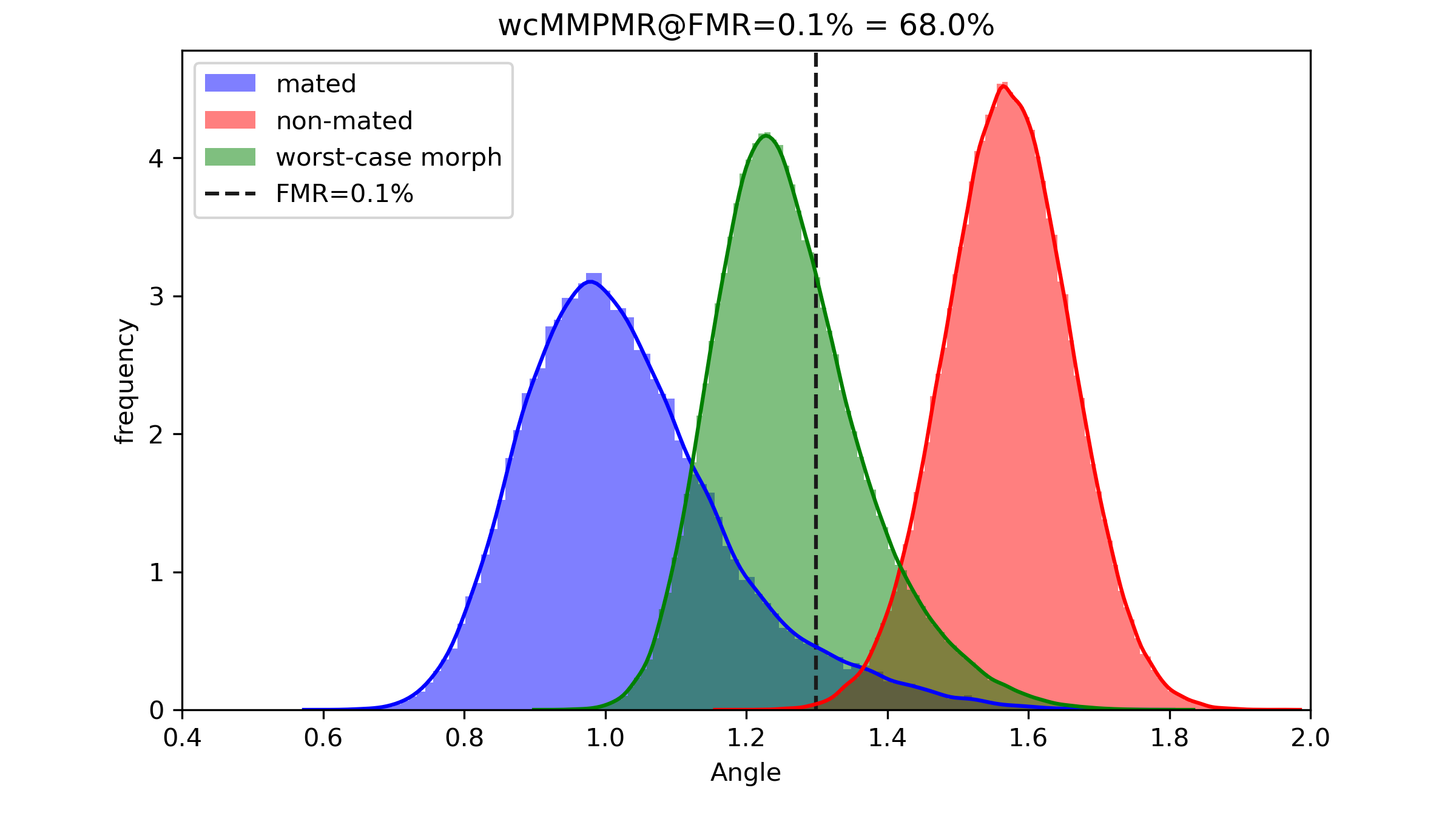}
	\includegraphics[width=0.45\textwidth]{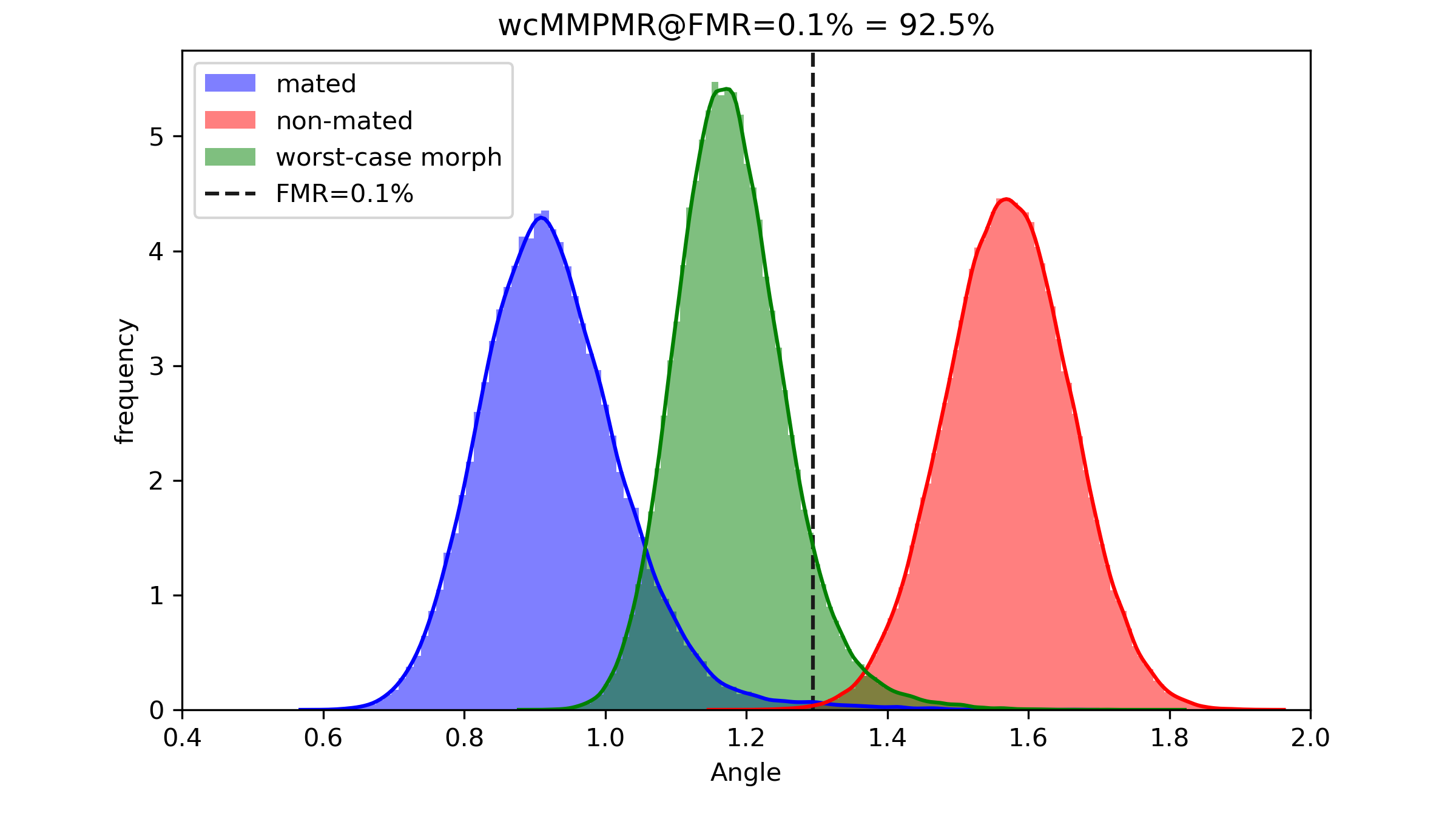}
	\includegraphics[width=0.45\textwidth]{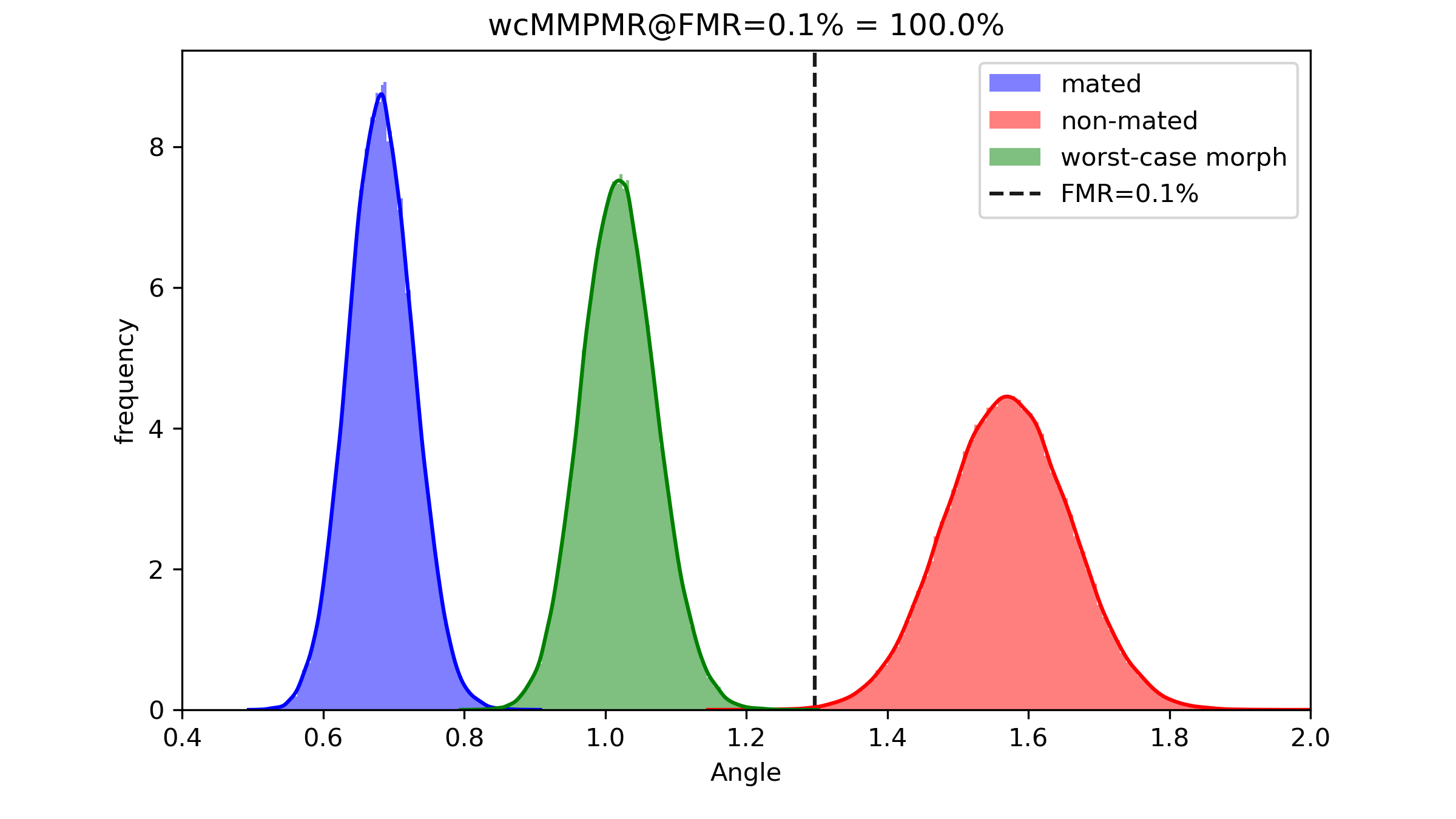}	
	\caption{Scores based on simulated embeddings using vMF with $\kappa$ drawn from $\mathcal{N}(\mu, \sigma)$. Top: $\mu=200$, $\sigma=60$, middle: $\mu=250$, $\sigma=50$, bottom: $\mu=500$, $\sigma=25$. These parameters were chosen to represent FR models with three levels of performance on normal images. The green distribution describes scores of comparisons between worst-case morph embeddings and probe embeddings. Because the threshold is set at $0.1\%$ FMR, the result of an improved mated distribution (i.e. with lower scores) is that more morphs are accepted.\label{fig:sim}}
\end{figure}

Using these sampled embeddings, we can calculate mated, non-mated and (worst-case) morph dissimilarity scores (angles). Then, we can determine error rates when using decision thresholds, for example at an FMR of 0.01\% or 0.1\%, see Fig.~\ref{fig:sim}. This figure explains why decision thresholds at e.g. $0.1\%$ FMR lead to improved FR systems being more vulnerable to morphing attacks, as was observed in \cite{GANmorphdetection, kelly2023worst, NIST_frvt}.

\subsection{Proposed metric}\label{sec:metrics}
As we explained in Section \ref{sec:sim}, setting an FR system's decision threshold such that the FMR is at most 0.1\% or 0.01\% on a given dataset inherently leads to high vulnerability to morphing attacks, especially for FR systems that perform well on normal data. Instead, in this section, we propose to set the decision threshold of an FR system such that the MMPMR calculated using worst-case embeddings is at most a fixed percentage~$x$. This provides the guarantee that no matter what new morphing algorithms are developed in the future, no more than $x\%$ of morphing attacks based on a given dataset can be successful. Mathematically, this threshold is defined as
\begin{equation}
t_{\text{wc}} := \underset{t}{\text{argmax}} \left\{ \text{wcMMPMR}(t) \leq x \right\},
\end{equation}
where $\text{wcMMPMR}(t)$ is the MMPMR at threshold $t$ calculated using worst-case embeddings. We call $\text{wcMMPMR}$ the \emph{worst-case MMPMR}. Fig. \ref{fig:fr_dmad} provides a visual explanation: the decision threshold is set using worst-case scores (as shown in the last row of the left-most column in Fig. \ref{fig:fr_dmad}a, \ref{fig:fr_dmad}b resp.).

While it is possible to evaluate D-MAD methods that rely on FR embeddings with worst-case morph embeddings, they do not provide a guarantee about the upper bound on the vulnerability to unknown morphing attacks like they do for FR systems. This is because the definition of worst-case embeddings relies on the underlying (dis)similarity score function used by an FR system, so we cannot guarantee that these cases are also worst-case for the D-MAD model.

\newcommand{\wiwide}{.25\linewidth}
\newcommand{\incwide}{\includegraphics[width=\wiwide]}
\begin{figure}[h]
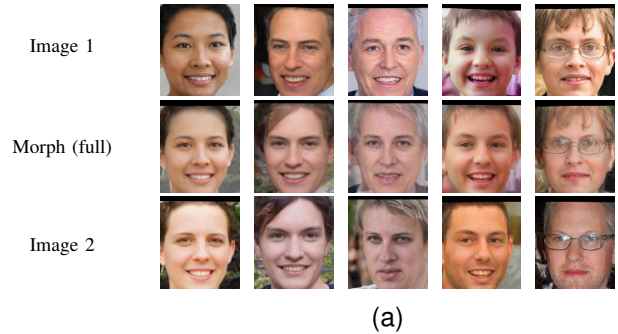

	\centering
	\resizebox{.1\textwidth}{!}{
		\begin{tabular}{c}			
			\raisebox{30pt}{Image 1}\\
			\raisebox{-10pt}{Morph (full)}\\
			\raisebox{-40pt}{Image 2}
		\end{tabular}
	}
	\subfloat[]{
		\resizebox{.35\textwidth}{!}{
			\begin{tabular}{ccccc}			
				\incwide{SynFace/001_00001.png}
				&\incwide{SynFace/001_00003.png}
				&\incwide{SynFace/001_00005.png}
				&\incwide{SynFace/001_00007.png}
				&\incwide{SynFace/001_00009.png} \\
				
				\incwide{SynFace/001_00001_001_00002_50.png}
				&\incwide{SynFace/001_00003_001_00004_50.png}
				&\incwide{SynFace/001_00005_001_00006_50.png}
				&\incwide{SynFace/001_00007_001_00008_50.png}
				&\incwide{SynFace/001_00009_001_00010_50.png} \\
				
				\incwide{SynFace/001_00002.png} 
				&\incwide{SynFace/001_00004.png}
				&\incwide{SynFace/001_00006.png}
				&\incwide{SynFace/001_00008.png}
				&\incwide{SynFace/001_00010.png}
			\end{tabular}
		}
	}
	\caption{Examples of morphs based on SynFace images. Top row: identity 1, middle row: landmark morph, bottom row: identity 2. \label{fig:SynFace_FRGC_pairs}}
\end{figure}

\subsection{Data}
To train the D-MAD approach based on FR embeddings, we generate synthetic identities using \cite{SynFace}, where we generate 1000 identities and 20 images per identity. We generate landmark morphs based on these identities: for each synthetic identity we choose the most similar identity using a pre-trained FR system \cite{ArcFace}, resulting in 816 identity pairs (if identity B is most similar to identity A, this does not guarantee that A is also most similar to B). There are 20 images per identity available, we compute one average embedding per identity
\begin{equation}
	z_{\text{avg}} = \frac{1}{20}\sum_{i=1}^{20} z_i,
\end{equation}
and select most similar identity pairs based on these averages. For each pair we generate 20 landmark morphs by morphing the first image of the first identity with the first image of the second identity, etc. This ensures that the bona fide and morphed training sets are balanced. See Fig. \ref{fig:SynFace_FRGC_pairs}.

\newcommand{\wi}{.2\linewidth}
\newcommand{\gap}{\hspace{-0.25cm}}
\newcommand{\inc}{\includegraphics[width=\wi]}
\begin{figure}[h]
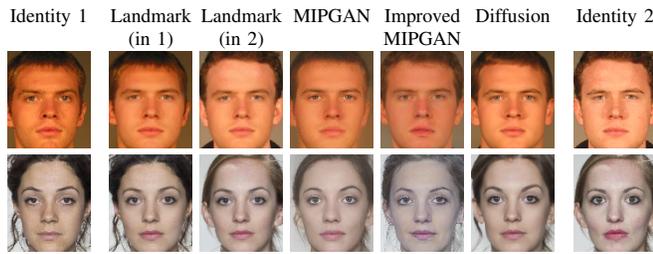

	\centering
	\large
	\resizebox{.5\textwidth}{!}{
		\begin{tabular}{ccccccc}
			Identity 1	& Landmark	& \gap Landmark	& \gap MIPGAN	& \gap Improved & \gap Diffusion 	& Identity 2	\\
			& (in 1) 	& \gap (in 2)  	& 				& \gap MIPGAN 	&  					&				\\
			
			\inc{morphs/04321d68}	&\inc{morphs/04321d68_04891d14_50_in0}	&\gap\inc{morphs/04321d68_04891d14_50_in1}	&\gap\inc{morphs/04321d68-vs-04891d14_mipgan}	&\gap\inc{morphs/04321d68_04891d14_imprmipgan}	&\gap\inc{morphs/04321d68_04891d14_diff}	&\inc{morphs/04891d14}		\\
			
			\inc{morphs/001_03}	&\inc{morphs/001_002}	&\gap\inc{morphs/002_001}	&\gap\inc{morphs/001_002_mipgan}	&\gap\inc{morphs/001_002_imprmipgan}	&\gap\inc{morphs/001_002_diff}	&\inc{morphs/002_03}
			
		\end{tabular}
	}
	\caption{\label{fig:morphs}Examples of morphs based on FRGC (top) and FRLL images (bottom). Landmark morphs are spliced into the background of contributor 1 or 2.}
\end{figure}

We evaluate the FR system and D-MAD approaches using images from \cite{FRGC} and \cite{FRLL}, and generate landmark morphs, MIPGAN morphs \cite{MIPGAN}, improved MIPGAN morphs \cite{kelly2023worst} and Diffusion morphs \cite{MorDiff}, see Fig. \ref{fig:morphs}.

\section{Conclusion}
Our experiments in Section \ref{sec:MADvsFR} showed that both face recognition (FR) and D-MAD models try to separate mated comparison scores from other comparison scores, regardless of whether they are morph or non-mated scores. Then, we explained that the vulnerability of FR systems is largely due to the way thresholds are selected and proposed the wcMMPMR as an alternative way to select thresholds. Specifically, a threshold can be set by choosing an acceptable upper bound for the vulnerability of an FR system to any type of morphing attack - even unknown ones. In future work, we would like to explore whether the vulnerability of FR systems can be reduced directly, rather than treating face recognition and morphing detection as separate problems. We recommend that whenever a D-MAD approach is evaluated, its ability to detect morphs should be compared to an FR system at an equivalent decision threshold. Using an FR system instead of D-MAD has the advantage that the wcMPPMR can be used to provide a guarantee about the vulnerability to morphing attacks of unknown types.


\bibliographystyle{IEEEtran}
\bibliography{draft}

\vfill

\end{document}